\documentclass{article}
\usepackage{ijcai25}

\usepackage{times}
\usepackage{soul}
\usepackage{url}
\usepackage[hidelinks]{hyperref}
\usepackage[utf8]{inputenc}
\usepackage[small]{caption}
\usepackage{graphicx}
\usepackage{amsmath}
\usepackage{amsthm}
\usepackage{booktabs}
\usepackage{algorithm}
\usepackage{algorithmic}
\usepackage[switch]{lineno}

\usepackage{multirow}
\usepackage{amsfonts}
\usepackage{bm} 
\usepackage{subcaption}

\title{Semantic-Guided Diffusion Model for Single-Step Image Super-Resolution}

\author{
Zihang Liu$^1$ \quad
Zhenyu Zhang$^2$ \quad
Hao Tang$^3$\thanks{Hao Tang is the corresponding author. This work was done while Zihang Liu was visiting Peking University.}
\affiliations
$^1$Beijing Institute of Technology\\
$^2$Nanjing University\\
$^3$School of Computer Science, Peking University\\
\emails
liuzihang@bit.edu.cn,
zhangjesse@foxmail.com,
haotang@pku.edu.cn
}

\begin{document}

\maketitle

\begin{abstract}
Diffusion-based image super-resolution (SR) methods have demonstrated remarkable performance. Recent advancements have introduced deterministic sampling processes that reduce inference from 15 iterative steps to a single step, thereby significantly improving the inference speed of existing diffusion models. However, their efficiency remains limited when handling complex semantic regions due to the single-step inference.
To address this limitation, we propose SAMSR, a semantic-guided diffusion framework that incorporates semantic segmentation masks into the sampling process. Specifically, we introduce the SAM-Noise Module, which refines Gaussian noise using segmentation masks to preserve spatial and semantic features. Furthermore, we develop a pixel-wise sampling strategy that dynamically adjusts the residual transfer rate and noise strength based on pixel-level semantic weights, prioritizing semantically rich regions during the diffusion process. To enhance model training, we also propose a semantic consistency loss, which aligns pixel-wise semantic weights between predictions and ground truth.
Extensive experiments on both real-world and synthetic datasets demonstrate that SAMSR significantly improves perceptual quality and detail recovery, particularly in semantically complex images. Our code is released at \href{https://github.com/Liu-Zihang/SAMSR}{https://github.com/Liu-Zihang/SAMSR}.
\end{abstract}

\section{Introduction}

\label{sec:intro}
Image super-resolution (SR) is a fundamental yet challenging problem in low-level vision, aiming to reconstruct a high-resolution (HR) image from a given low-resolution (LR) input \cite{wang2020deep}. The task is inherently ill-posed due to the complexity and unknown nature of degradation models in real-world scenarios. Recently, diffusion models, as an emerging generative paradigm, have achieved unprecedented success in image generation and demonstrated remarkable potential in various low-level vision tasks, including image editing, inpainting, and colorization.

\begin{figure}[t]
    \centering
    \begin{subfigure}[b]{0.482\textwidth}
        \centering
        \includegraphics[
          width=\textwidth,
          trim=0 80 0 50, clip  
        ]{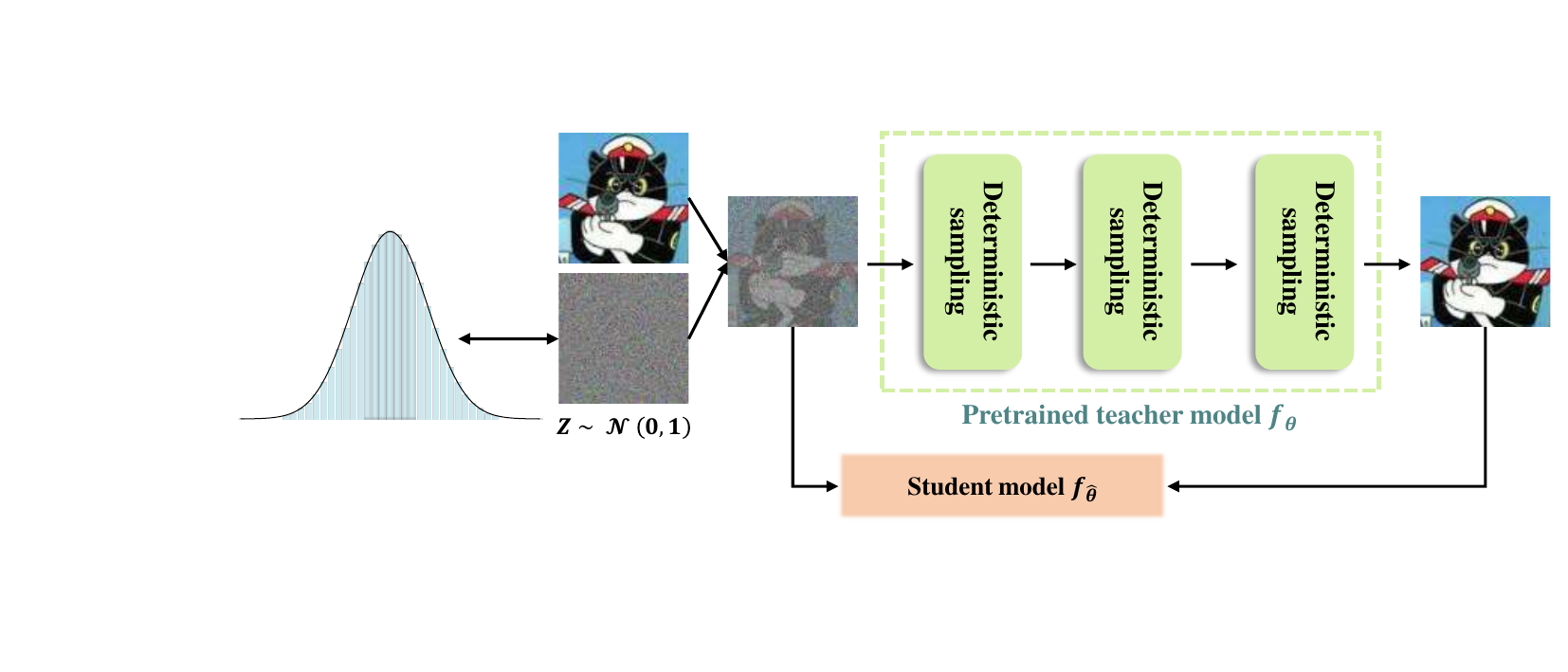}
        \caption{The SOTA method SinSR shortens the Markov chain to speed up the inference process by introducing the deterministic sampling strategy.}
        \label{fig:original_noise}
    \end{subfigure}
    \hfill
    \begin{subfigure}[b]{0.482\textwidth}
        \centering
        \includegraphics[width=\textwidth]{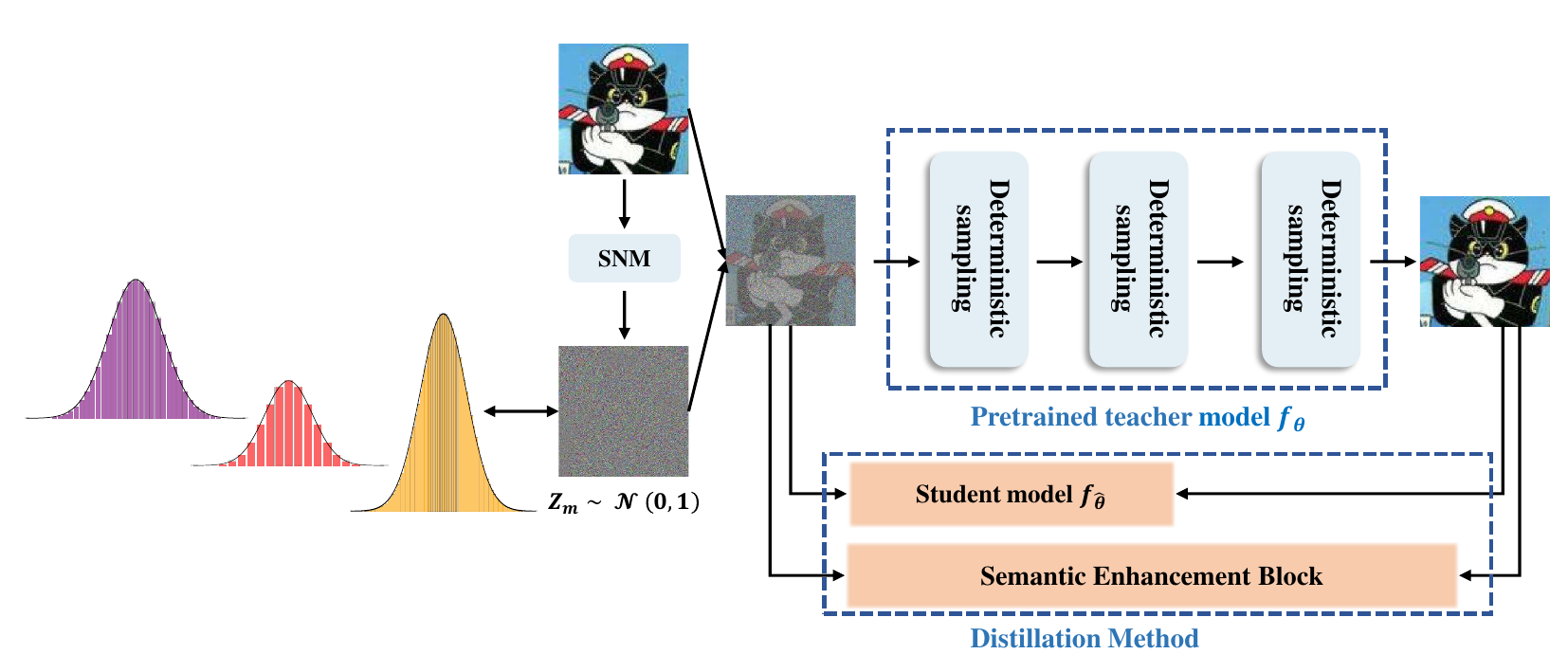}
        \caption{A simplified pipeline of the proposed method SAMSR. It refines Gaussian noise and distillation method using segmentation masks to preserve spatial and semantic features.}
        \label{fig:sam_diffsr_noise}
    \end{subfigure}
    
    \caption{A comparison between the most recent SOTA one-step SR model and our SAMSR model. Different from recent works with simple noise distribution, the proposed method incorporates semantic segmentation information into the noise distribution and gause diffusion process.} 
    \label{fig:noise_distribution_comparison}
    \vspace{-0.4cm}
\end{figure}

Currently, strategies for employing diffusion models in image SR can be broadly categorized into two approaches: (i) inserting the LR image as input to the denoiser and retraining the model from scratch, \cite{rombach2022high,saharia2022image} and (ii) utilizing an unconditional pre-trained diffusion model as a prior and modifying its reverse path to generate the desired HR image \cite{choi2021ilvr,chung2022come,wang2021real}. However, both strategies face significant computational efficiency challenges. Conventional methods typically initiate from pure Gaussian noise, failing to leverage the prior knowledge embedded in the LR image, consequently requiring a substantial number of inference steps to achieve satisfactory performance and severely constraining the practical application of diffusion-based SR techniques.

Although various acceleration techniques have been proposed to accelerate diffusion model sampling \cite{lu2022dpm,lugmayr2020srflow,song2020denoising}, these methods often compromise performance in low-level vision domains that demand high fidelity. Recent innovative research has begun to reformulate the diffusion process in image restoration tasks, attempting to model the initial step as a combination of LR images and random noise \cite{yue2024resshift}. However, the inference speed remains limited. Some subsequent works have explored deterministic sampling strategies for image SR, learning bidirectional deterministic mappings between noise and HR image generation to improve inference speed \cite{wang2024sinsr}, which is shown in Fig.~\ref{fig:original_noise}. However, these models frequently suffer from limited authenticity and reduced capability in processing complex semantic images due to constrained inference steps.

To address these challenges, we propose a semantic segmentation-based pixel-wise sampling framework, SAMSR as shown in Fig.~\ref{fig:sam_diffsr_noise}, a semantic segmentation guided framework to address the limitations of deterministic sampling in diffusion-based image SR. Existing methods often apply uniform noise addition and global parameters, making it challenging to recover fine details in semantically complex regions. To overcome this, we introduce the SAM-Noise Module, which leverages segmentation masks generated by the Segment Anything Model (SAM) to perform spatially adaptive noise sampling, preserving both semantic and spatial features. Additionally, we propose a semantic-guided forward process that dynamically adjusts the residual transfer rate and noise strength at the pixel level based on semantic weights, enabling prioritized recovery of semantically rich regions. To enhance training, a semantic consistency loss is introduced to align the pixel-wise semantic weights between the prediction and the ground truth. These innovations allow SAMSR to effectively utilize semantic information, achieving superior performance in both real-world and synthetic datasets, particularly in recovering fine details and textures.

Our main contributions are summarized as follows:
(i) We introduce, for the first time, a segmentation-mask-based random noise sampling method. This approach performs single-distribution noise sampling separately within each masked region and normalizes and combines them, allowing the Gaussian noise to retain both the spatial and semantic features of the original image.
(ii) We leverage the segmentation masks obtained from SAM to derive pixel-level sampling hyperparameters, differentiating the noise addition speed for pixels with varying levels of semantic richness. This ensures that semantically rich regions are distinctly recovered within a single sampling step.
(iii) We propose a novel consistency semantic loss that utilizes ground-truth images during training to enhance the model's understanding and application of region segmentation masks, leading to improved performance.

\section{Related Work}

\begin{figure*}[t]
    \centering
    \begin{minipage}[t]{0.4875\textwidth}
        \centering
        \includegraphics[width=\textwidth]{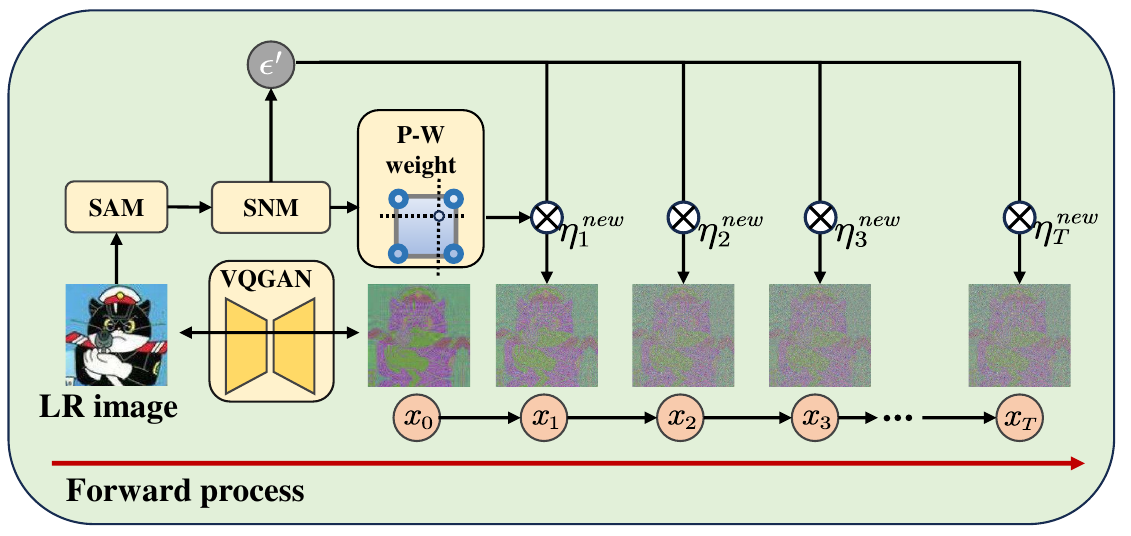}
        \caption*{(a) Forward process}
        \label{fig:(a) Forward process}
    \end{minipage}%
    \hspace{0.0025\textwidth}
    \hspace{0.0025\textwidth}%
    \begin{minipage}[t]{0.4975\textwidth}
        \centering
        \includegraphics[width=\textwidth]{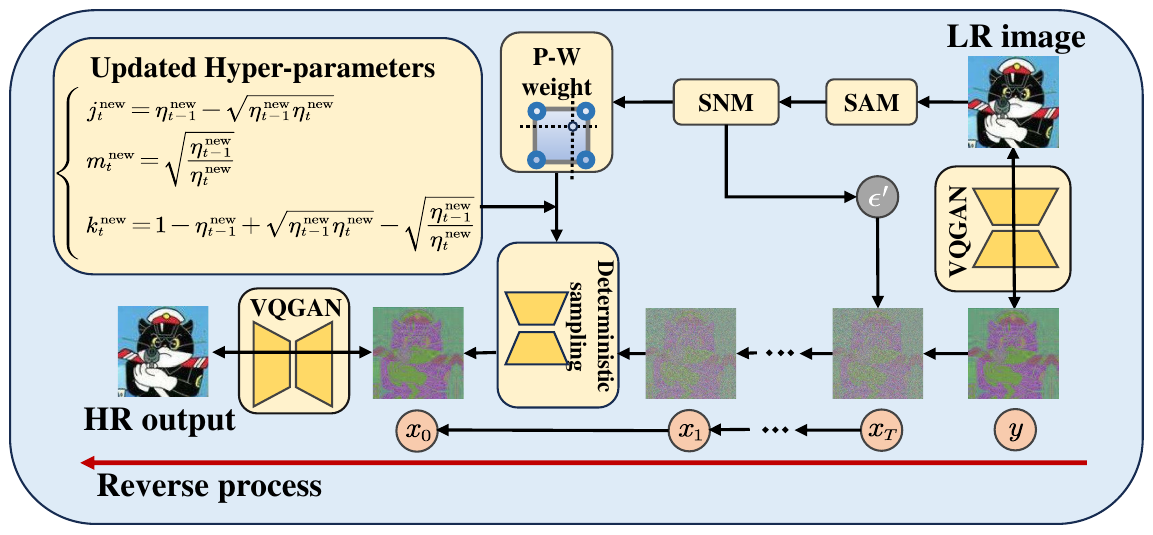}
        \caption*{(b) Reverse process}
        \label{fig:(b) Reverse process}
    \end{minipage}
    \caption{The overall framework of the proposed SAMSR method. The SAM-Noise Module (SNM) generates semantically-guided noise maps $\epsilon'$ for the forward process while computing pixel-wise semantic weights. These weights are utilized to adaptively adjust the residual transfer rate and noise strength in both forward and reverse processes, enabling fine-grained control over semantic region reconstruction.}
    \label{fig:framework}
    \vspace{-0.2cm}
\end{figure*}

\noindent\textbf{Advances in Super-Resolution Techniques.}
Super-resolution (SR) has undergone significant evolution, transitioning from early methods based on handcrafted priors to deep learning-based approaches. Early SR algorithms leveraged priors such as non-local similarity \cite{sun2008image}, sparse coding \cite{yang2010image,Cai_2019_ICCV}, and low-rankness \cite{milanfar2012tour,Cai_2019_ICCV}. These handcrafted approaches were effective for basic degradations but lacked flexibility and generalization for complex real-world scenarios \cite{zhang2021designing}.

Deep learning revolutionized SR with the introduction of convolutional neural networks (CNNs) in SRCNN \cite{dong2014learning}, which marked the beginning of a series of innovations. Residual learning \cite{zhang2018residual}, attention mechanisms and transformers further improved SR performance in terms of fidelity and perceptual quality \cite{cao2023ciaosr}. Generative adversarial networks (GANs) like ESRGAN \cite{wang2018esrgan} pushed SR towards generating more realistic textures but often introduced perceptual artifacts and training instability \cite{ledig2017photo}.

Recently, diffusion models, initially developed for generative tasks \cite{saharia2022image}, have emerged as promising tools for SR. Unlike GANs, diffusion models iteratively refine Gaussian noise into structured HR images, offering superior theoretical guarantees and perceptual quality. Methods like SR3 \cite{saharia2022image} and SRDiff \cite{song2020denoising} adapt diffusion models for SR by modifying their reverse processes or combining LR images with noise. However, these methods are often computationally expensive, requiring hundreds or thousands of iterative steps \cite{ho2020denoising,lu2022dpm,lugmayr2020srflow}.

\noindent\textbf{Acceleration of Diffusion Models for SR.}
The inefficiency of diffusion models has motivated extensive research into acceleration techniques. DDIM introduced deterministic sampling paths, which reduced inference steps but often compromised image fidelity in SR tasks \cite{ho2020denoising,lu2022dpm}. Progressive sampling methods, like those used in Latent Diffusion Models (LDM), offered more efficient sampling but still required tens of steps to achieve satisfactory results \cite{rombach2022high}.

To address these challenges, ResShift \cite{yue2024resshift} proposed embedding LR information directly into the Markov chain, significantly reducing sampling steps while preserving fidelity. Similarly, SinSR \cite{wang2024sinsr} distilled the mapping between Gaussian noise and HR images into a lightweight student network, achieving single-step SR with up to a tenfold speedup. However, these methods face limitations in semantically complex regions, as they rely on uniform noise priors rather than adaptive spatial information \cite{yue2024resshift,wang2024sinsr}.
Recent advancements such as HoliSDiP \cite{tsao2024holisdip} integrate semantic segmentation with diffusion frameworks, providing global and localized semantic information for better spatial fidelity. This approach demonstrates the potential to further reduce sampling complexity while enhancing image quality by leveraging holistic semantic priors \cite{cao2023ciaosr,tsao2024holisdip}.

\noindent\textbf{Applications of SAM in Various Domains.}
Segment Anything Model (SAM) \cite{kirillov2023segment} and its successor SAM2 \cite{ravi2024sam} have been adapted to diverse fields, including medical imaging, video analysis, and super-resolution. Below, we summarize its applications in these areas and their relevance to this work.
\underline{(i) SAM in Medical Image Segmentation:} SAM has been widely adopted in medical imaging for tasks like tumor segmentation and organ delineation. Models like Med-SAM-Adapter and SAMed leverage parameter-efficient fine-tuning strategies, such as adapters and low-rank adaptations, to tailor SAM for domain-specific applications \cite{wu2023medical,zhang2023customized,leng2024self}. For instance, SAM-based segmentation has been explored in retinal vessel segmentation \cite{zhang2024light} and polyp segmentation \cite{xiong2024sam2}. Additionally, MedSAM-2 \cite{zhu2024medical} demonstrates how memory mechanisms can adapt SAM for 3D segmentation tasks, allowing it to handle unordered medical image slices. These works emphasize SAM's ability to address challenges like data scarcity and fine-grained segmentation in medical domains \cite{shen2024interactive,yao2023false}.
\underline{(ii) SAM in Video Understanding and Analysis:} In video segmentation and tracking, SAM has been adapted to dynamic contexts by incorporating temporal modeling and memory management. For example, SAMURAI introduces motion-aware memory mechanisms to enhance object tracking under occlusion and rapid motion \cite{yang2024samurai}. Similarly, SAM2Long uses tree-based memory architectures to improve segmentation consistency across long video sequences \cite{ding2024sam2long}. In surgical video segmentation, Surgical SAM2 achieves real-time segmentation by employing efficient frame pruning, reducing computational demands while maintaining accuracy \cite{liu2024surgical}. These adaptations enable SAM to excel in spatiotemporal tasks requiring consistent object identity across frames.
\underline{(iii) SAM in Image and Video SR:} SAM's semantic capabilities have recently been extended to super-resolution tasks. SAM Boost utilizes semantic priors to improve alignment and fusion in video super-resolution, enabling better handling of large motions and occlusions \cite{lu2023can,liu2024explainable}. HoliSDiP combines SAM-derived segmentation maps with diffusion models, providing spatial guidance for improved image super-resolution in real-world scenarios \cite{tsao2024holisdip}. By integrating SAM's zero-shot segmentation and spatial adaptability, these works demonstrate the potential of SAM in enhancing detail reconstruction and semantic consistency in SR tasks.

\section{Methodology}
\label{sec:Methodology}

\subsection{Overview}

The SinSR model and the ResShift model differ primarily in their ability to reduce the number of inference steps from 15 to a single step through a deterministic sampling strategy. In the original SinSR, the forward diffusion process begins by combining a low-resolution (LR) image \( y \) with Gaussian noise \( \epsilon \), scaled by a residual transfer rate \( \eta_t \) and noise strength \( \kappa \). This is formulated as:
\begin{equation}
q(x_t | x_0, y) = N(x_t; x_0 + \eta_t (y - x_0), \kappa^2 \eta_t I),
\end{equation}
while the reverse process is represented by a deterministic mapping:
\begin{equation}
x_{t-1} = k_t \hat{x}_0 + m_t x_t + j_t y,
\end{equation}
where \( k_t \), \( m_t \), and \( j_t \) are coefficients derived from \( \eta_t \).
Although SinSR achieves single-step sampling efficiency, its reliance on global Gaussian noise \( \epsilon \) and uniform diffusion parameters \( \eta_t \) and \( \kappa \) limits its flexibility in handling semantically complex regions. 

To address this, we propose the \textbf{SAMSR} framework, which introduces semantic guidance via the SAM-Noise Module and SAM-based Forward Process. The SAM-Noise Module refines the global Gaussian noise \( \epsilon \) into a spatially adaptive noise map $\epsilon'$, as described in Sec.~\ref{sec:sam-noise}. The SAM-based Forward Process dynamically adjusts the residual transfer rate \( \eta_t \) and noise strength \( \kappa \) based on semantic weights derived from SAM masks, as detailed in Sec.~\ref{sec:sam-forward}. Furthermore, to enhance the model's understanding of semantic information during training, we introduce a semantic consistency loss that aligns semantic features between predictions and ground truth, which will be elaborated in Sec.~\ref{sec:Semantic Consistency Loss}.

The overall framework of SAMSR is illustrated in Fig.~\ref{fig:framework}. As shown in Fig. 2(a), the forward process combines the SAM-Noise Module and pixel-wise weight computation to generate semantically-guided noise. In the reverse process (Fig. 2(b)), these semantic cues are utilized to dynamically adjust the sampling hyperparameters, enabling region-aware image reconstruction. This semantic-guided framework allows SAMSR to better preserve details in semantically rich regions while maintaining global consistency.

\subsection{SAM-Noise Module}
\label{sec:sam-noise}

\begin{figure}[h]
    \centering
    \includegraphics[width=1\linewidth]{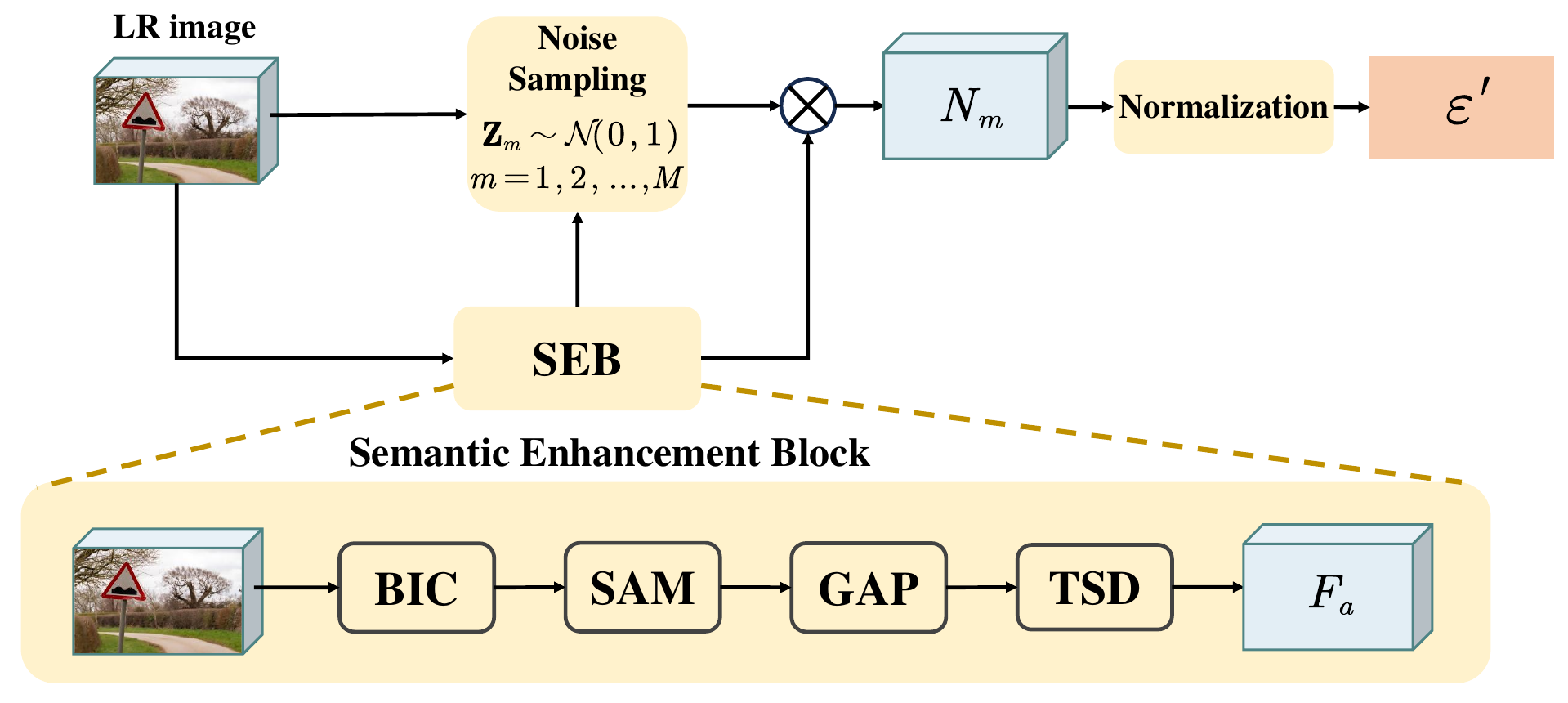}
    \caption{Architecture of the SAM-Noise Module. The module consists of two main components: (1) A Semantic Enhancement Block integrating bicubic interpolation (BIC), Segment Anything Model (SAM), global average pooling (GAP), and thresholding (TSD) operations for mask generation; (2) A noise sampling and normalization pipeline that leverages semantic information to produce spatially-adaptive noise distributions. This design enables semantically-guided noise generation while preserving structural consistency.}
    \label{fig:sam_noise_module}
\end{figure}

The SAM-Noise Module aims to integrate the spatial and semantic features of the original image into Gaussian noise, thereby enhancing the diffusion model's ability to handle images with complex semantic information. As shown in Fig. \ref{fig:sam_noise_module}, to improve the accuracy of semantic segmentation, the given LR image is first passed through a bicubic interpolation process as input to the SAM. The resulting mask information is then processed through global average pooling and thresholding operations to obtain a binary tensor mask of the original LR image. The specific computation formula is as follows:
\begin{equation}
\begin{aligned}
    F_p &= BIC(LR), \quad F_p \in \mathbb{R}^{3 \times 4H \times 4W}, \\
    F_d &= SAM(F_p), \quad F_d \in \mathbb{R}^{M \times 4H \times 4W}, \\
    F_u &= GAP(F_d), \quad F_u \in \mathbb{R}^{M \times H \times W}, \\
    F_a &= TSD(F_u), \quad F_a \in \mathbb{R}^{M \times H \times W}.
\end{aligned}
\end{equation}
where $BIC$ is the bicubic interpolation, the $SAM$ is the segment anything model, and the $GAP$  is the global average pooling.
To obtain the final binary mask, a thresholding operation is applied to $F_u$, where values greater than a predefined threshold  $T$ are set to $1$, and the rest are set to $0$:
\begin{equation}
F_a(i, j) =
\begin{cases}
1, & \text{if } F_u(i, j) > T, \\
0, & \text{otherwise}.
\end{cases}
\end{equation}
We set the threshold $T$ to $0.5$.
To generate refined noise for the forward diffusion process, we incorporate the binary mask \( F_a \) into the noise sampling procedure. Specifically, we first sample \( M \) independent noise maps \( \mathbf{Z}_m \in \mathbb{R}^{3\times H \times W} \) from a standard normal distribution \( \mathcal{N}(0, 1) \), where \( m = 1, 2, \dots, M \). These noise maps are then multiplied by the binary mask in the element \( F_a \), ensuring that the noise is applied only within the regions covered by the mask. Mathematically, the masked noise \( \mathbf{N}_m \) is defined as:
\begin{equation}
\mathbf{N}_m = F_a \odot \mathbf{Z}_m, \quad \mathbf{Z}_m \sim \mathcal{N}(0, 1), \; m = 1, 2, \dots, M,
\end{equation}
where \( \odot \) denotes the element-wise multiplication.
Next, the \( M \) masked noise maps are summed to produce a combined noise map \( \mathbf{N}_{\text{sum}} \):
\begin{equation}
\mathbf{N}_{\text{sum}} = \sum_{m=1}^M \mathbf{N}_m.
\end{equation}

To ensure the noise map is normalized for the diffusion process, we standardize \( \mathbf{N}_{\text{sum}} \) to have zero mean and unit variance. The final noise map $\epsilon'$ is computed as follows:

\begin{equation}
\mathbf{\epsilon'} = \frac{\mathbf{N}_{\text{sum}} - \mu_{\mathbf{N}_{\text{sum}}}}{\sigma_{\mathbf{N}_{\text{sum}}}},
\end{equation}
where \( \mu_{\mathbf{N}_{\text{sum}}} \) and \( \sigma_{\mathbf{N}_{\text{sum}}} \) represent the mean and standard deviation of \( \mathbf{N}_{\text{sum}} \).
The resulting noise $\epsilon'$ is used as the input to the forward diffusion process, ensuring that noise is spatially restricted to the mask-covered regions while maintaining a normalized distribution.

\subsection{SAM-based Forward and Reverse Process}
\label{sec:sam-forward}

To refine the residual transfer rate and noise strength based on semantic regions, we introduce a pixel-wise weight matrix \( W(x, y) \), which is derived from the binary masks \( F_a \). Specifically, for each pixel location \( (x, y) \), \( W(x, y) \) is defined as the normalized coverage across all \( M \) masks, where normalization is performed using the maximum pixel coverage:

\begin{equation}
    W(x, y) = \frac{\sum_{m=1}^M F_a^m(x, y)}{\max_{(x', y')} \sum_{m=1}^M F_a^m(x', y')}, \quad W(x, y) \in [0, 1],
    \label{eq:W_(x,y)}
\end{equation}
where \( F_a^m(x, y) \in \{0, 1\} \) represents the binary value of the \( m \)-th mask at pixel location \( (x, y) \), and \( M \) denotes the total number of masks. The denominator represents the maximum coverage among all pixels in the image.

Using the weight matrix \( W(x, y) \), the residual transfer rate \( \eta_t \) and noise strength \( \kappa \) are adjusted as follows:

\begin{equation}
\label{Eq:eta_new and kappa_new}
\begin{aligned}
    \eta_t^{\text{new}}(x, y) &= \eta_t \cdot \left( 1 + m \cdot W(x, y) \right), \\
    \kappa^{\text{new}}(x, y) &= \kappa \cdot \left( 1 - m \cdot W(x, y) \right),
\end{aligned}
\end{equation}
where $m$ is the hyper-parameter that controls the noise addition speed and intensity for pixels with different levels of semantic richness during the forward diffusion process.

Utilizing the Pixel-wise Weight Matrix, we update the forward and reverse process to introduce semantic adaptiveness into the deterministic sampling framework. Specifically, the adjustments are applied to the residual transfer rate \( \eta_t \) and noise strength \( \kappa \), enabling pixel-wise control based on semantic guidance.

Therefore, using the noise map $\epsilon'$, we can update the forward process of the diffusion model starting from an initial state from the LR image $y$ as below:
\begin{equation}
    x_T = y + \kappa^{\text{new}} \sqrt{\eta_t^{\text{new}}}\mathbf{\epsilon'}.
\end{equation}

The updated reverse process at time step \( t \) is given as:
\begin{equation}
x_{t-1}(x, y) = k_t^{\text{new}} \hat{x}_0(x, y) + m_t^{\text{new}} x_t(x, y) + j_t^{\text{new}} y(x, y),
\end{equation}
where \( k_t^{\text{new}} \), \( m_t^{\text{new}} \), and \( j_t^{\text{new}} \) are the updated coefficients derived from the pixel-wise adjusted residual transfer rate \( \eta_t^{\text{new}} \). These parameters are defined as:
\begin{equation}
\begin{aligned}
k_t^{\text{new}} &= 1 - \eta_{t-1}^{\text{new}} + \sqrt{\eta_{t-1}^{\text{new}} \eta_t^{\text{new}}} - \sqrt{\frac{\eta_{t-1}^{\text{new}}}{\eta_t^{\text{new}}}}, \\
m_t^{\text{new}} &= \sqrt{\frac{\eta_{t-1}^{\text{new}}}{\eta_t^{\text{new}}}}, \\
j_t^{\text{new}} &= \eta_{t-1}^{\text{new}} - \sqrt{\eta_{t-1}^{\text{new}} \eta_t^{\text{new}}},
\end{aligned}
\label{eq:k,m,j}
\end{equation}
where \( \eta_t^{\text{new}} \) represents the dynamically adjusted residual transfer rate at each pixel location \((x, y)\), which is computed in Eq. \eqref{Eq:eta_new and kappa_new}.

\begin{algorithm}[t]
\caption{Training the Pixel-wise Sampling Framework}
\label{alg:training}
\begin{algorithmic}[1]
\REQUIRE Pre-trained teacher diffusion model \( f_\theta \)
\REQUIRE Paired training set \( (X, Y) \)
\STATE Init \( f_{\hat{\theta}} \) from the pre-trained model, \textit{i.e.,} \( \hat{\theta} \gets \theta \)
\WHILE{not converged}
    \STATE Sample \( x_0, y \sim (X, Y) \)
    \STATE Compute \( W_{y}(x, y) \) using Equation~\ref{eq:W_(x,y)}
    \STATE Compute \( \kappa ^{new}, \eta_T^{\text{new}} \) using Equation~\ref{Eq:eta_new and kappa_new}
    \STATE Compute \(  k_t^{\text{new}}, m_t^{\text{new}}, j_t^{\text{new}} \) using Equation~\ref{eq:k,m,j}
    \STATE Sample \( \epsilon \sim \mathcal{N}(0, \left( \kappa ^{new} \right) ^2 \eta_T^{\text{new}} \mathbf{I}) \)
    \STATE \( x_T = y + \epsilon \)
    \FOR{\( t = T, T-1, \dots, 1 \)}
        \IF{\( t = 1 \)}
            \STATE \( \hat{x}_0 = f_\theta(x_1, y, 1) \)
        \ELSE
            \STATE \( x_{t-1} = k_t^{\text{new}} f_\theta(x_t, y, t) + m_t^{\text{new}} x_t + j_t^{\text{new}} y \)
        \ENDIF
    \ENDFOR
    \STATE \( \mathcal{L}_{\text{distill}} = L_{\text{MSE}}(f_{\hat{\theta}}(x_T, y, T), \hat{x}_0) \)
    \STATE \( \mathcal{L}_{\text{inverse}} = L_{\text{MSE}}(f_{\hat{\theta}}(\hat{x}_0, y, 0), x_T) \)
    \STATE \( \hat{x}_T = f_{\hat{\theta}}(x_0, y, 0) \)
    \STATE \( \mathcal{L}_{\text{gt}} = L_{\text{MSE}}(f_{\hat{\theta}}(\text{detach}(\hat{x}_T), y, T), x_0) \)
    \STATE Compute \( W_{\hat{x}_0}(x, y), W_{x_0}(x, y)\) using Equation~\ref{eq:W_(x,y)}
    \STATE \( \mathcal{L}_{\text{SC}} = L_{\text{MSE}}(W_{\hat{x}_0}(x, y), W_{x_0}(x, y)) \)
    \STATE \( \mathcal{L} = \mathcal{L}_{\text{distill}} + \mathcal{L}_{\text{inverse}} + \mathcal{L}_{\text{gt}} +\lambda\mathcal{L}_{\text{SC}}\)
    \STATE Perform a gradient descent step on \( \nabla_{\hat{\theta}} \mathcal{L} \)
\ENDWHILE
\RETURN The student model \( f_{\hat{\theta}} \)
\end{algorithmic}
\label{al_1}
\end{algorithm} 

\subsection{Semantic Consistency Loss}
\label{sec:Semantic Consistency Loss}

To further incorporate semantic guidance into the training process, we introduce a Semantic Consistency Loss \( L_{\text{SC}} \), which aligns the semantic weights between the predicted output \( \hat{x}_0 \) and the ground truth \( x_0 \). Using the Pixel-wise Weight Matrix \( W(x, y) \) defined in Sec.~\ref{sec:sam-forward}, we compute the normalized semantic weights \( W_{\hat{x}_0}(x, y) \) for the predicted image and \( W_{x_0}(x, y) \) for the ground truth. The loss is formulated as:
\begin{equation}
\mathcal{L}_{\text{SC}} = L_{\text{MSE}}\left( W_{\hat{x}_0}(x, y), W_{x_0}(x, y) \right).
\end{equation}

This additional loss is integrated into the original training objective, which consists of the distillation loss \( \mathcal{L}_{\text{distill}} \), reverse loss \( \mathcal{L}_{\text{reverse}} \), and ground truth loss \( \mathcal{L}_{\text{gt}} \). The updated training objective is defined as:
\begin{equation}
\hat{\theta} = \arg \min_{\hat{\theta}} \mathbb{E}_{y, x_0, x_T} \left[ \mathcal{L}_{\text{distill}} + \mathcal{L}_{\text{reverse}} + \mathcal{L}_{\text{gt}} + \lambda \mathcal{L}_{\text{SC}}\right],
\end{equation}
where \( \lambda \) is a hyper-parameter that controls the contribution of the semantic consistency loss.

By explicitly enforcing alignment between the semantic weights of the prediction and ground truth, the proposed \( L_{\text{SC}} \) improves the model's ability to utilize semantic segmentation masks effectively, leading to better semantic understanding during training. The overall of the
proposed method is summarized in Algorithm \ref{al_1}.

\section{Experiments}
\label{sec:Experiment}

\subsection{Experimental Setup}

\begin{figure*}[t!]
    \centering
    \scriptsize  
    \begin{tabular}{c@{\hspace{2mm}}c@{\hspace{2mm}}c@{\hspace{2mm}}c@{\hspace{2mm}}c@{\hspace{2mm}}c}
        \begin{tabular}{c}
            \raisebox{-0.595\height}[0pt][0pt]{\includegraphics[width=0.275\textwidth]{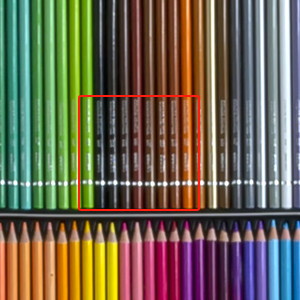}} \\
            \raisebox{-17.25\height}[0pt][0pt]{(a) LR input}
        \end{tabular} &
        \includegraphics[width=0.12\textwidth]{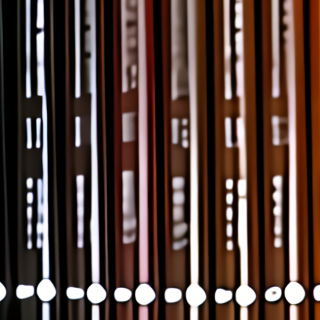} &
        \includegraphics[width=0.12\textwidth]{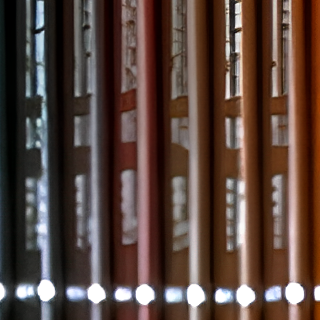} &
        \includegraphics[width=0.12\textwidth]{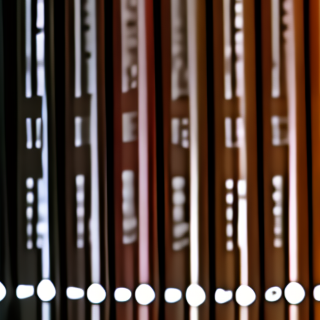} &
        \includegraphics[width=0.12\textwidth]{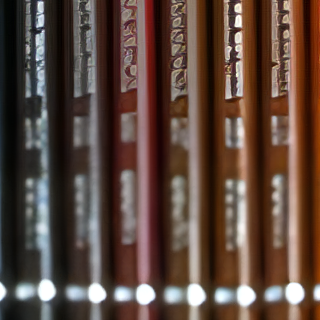} &
        \includegraphics[width=0.12\textwidth]{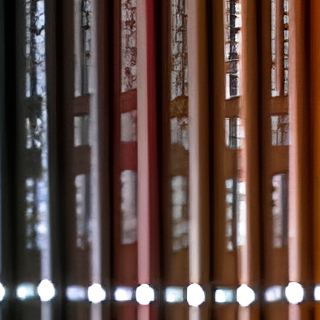} \\[-1mm]
        & (b) ESRGAN & (c) SwinIR & (d) DASR & (e) BSRGAN & (f) RealESRGAN \\[2mm]
        & \includegraphics[width=0.12\textwidth]{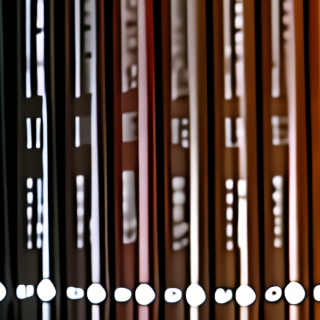} &
        \includegraphics[width=0.12\textwidth]{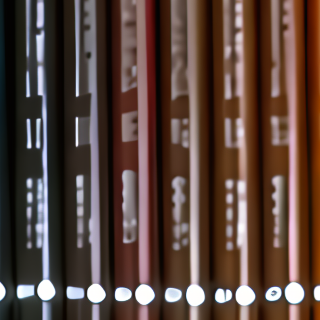} &
        \includegraphics[width=0.12\textwidth]{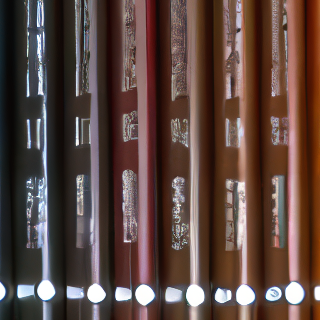} &
        \includegraphics[width=0.12\textwidth]{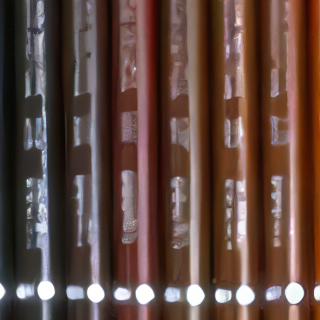} &
        \includegraphics[width=0.12\textwidth]{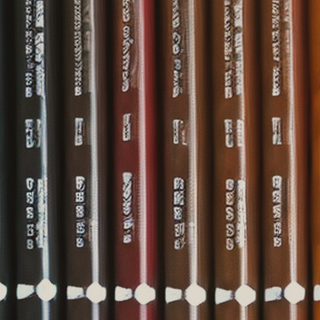} \\[-1mm]
        & \raisebox{-0.75\height}[0pt][0pt]{ (g) LDM-30} & \raisebox{-0.75\height}[0pt][0pt]{ (h) LDM-100} & \raisebox{-0.75\height}[0pt][0pt]{ (i) ResShift-15} & \raisebox{-0.75\height}[0pt][0pt]{ (j) SinSR} & \raisebox{-0.75\height}[0pt][0pt]{ (k) SAMSR} \\[2mm]
    \end{tabular}
    \begin{tabular}{c@{\hspace{2mm}}c@{\hspace{2mm}}c@{\hspace{2mm}}c@{\hspace{2mm}}c@{\hspace{2mm}}c}
        \begin{tabular}{c}
            \raisebox{-0.595\height}[0pt][0pt]{\includegraphics[width=0.275\textwidth]{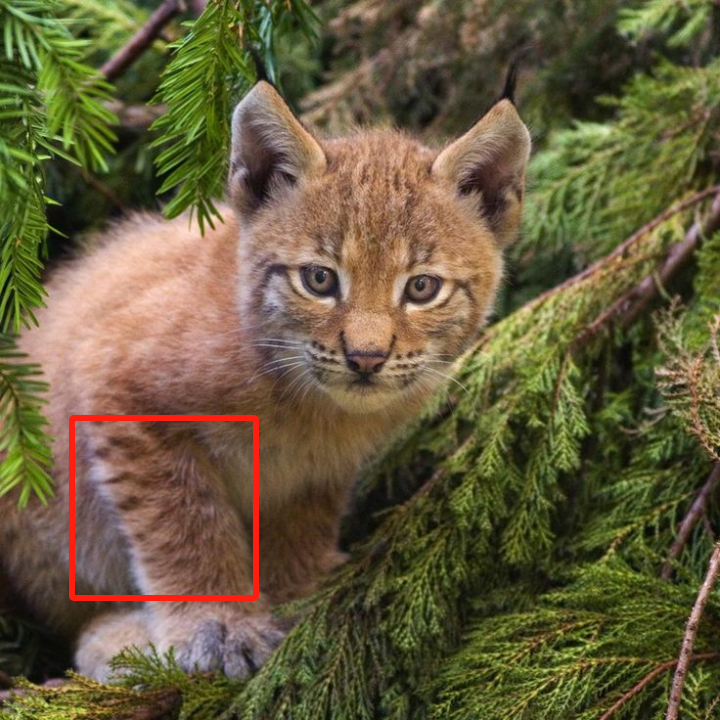}} \\
            \raisebox{-17.25\height}[0pt][0pt]{(a) LR input}
        \end{tabular} &
        \includegraphics[width=0.12\textwidth]{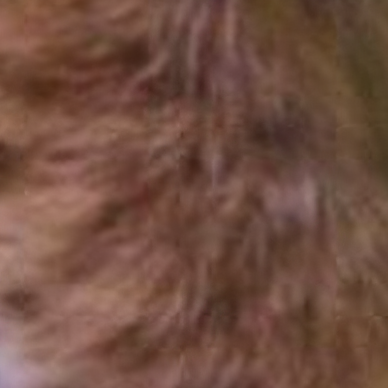} &
        \includegraphics[width=0.12\textwidth]{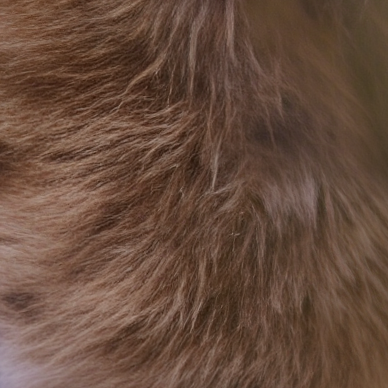} &
        \includegraphics[width=0.12\textwidth]{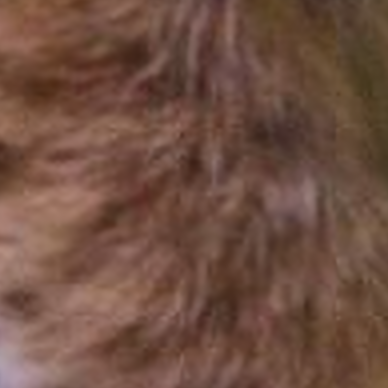} &
        \includegraphics[width=0.12\textwidth]{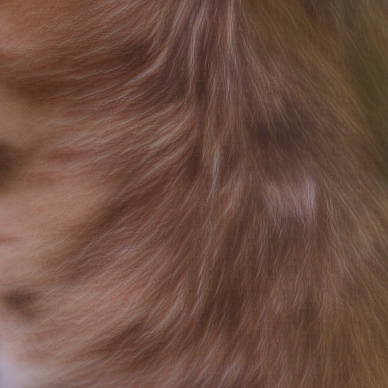} &
        \includegraphics[width=0.12\textwidth]{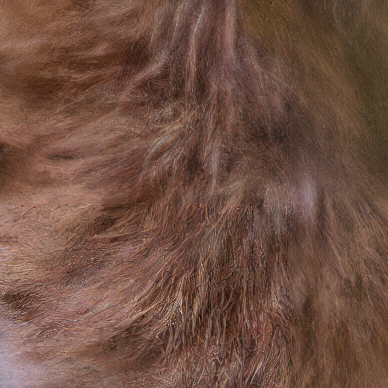} \\[-1mm]
        & (b) ESRGAN & (c) SwinIR & (d) DASR & (e) BSRGAN & (f) RealESRGAN \\[2mm]
        & \includegraphics[width=0.12\textwidth]{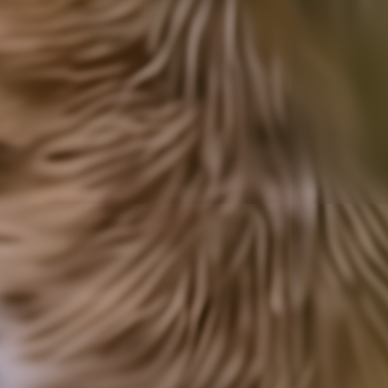} &
        \includegraphics[width=0.12\textwidth]{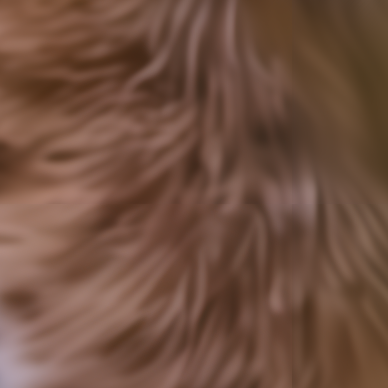} &
        \includegraphics[width=0.12\textwidth]{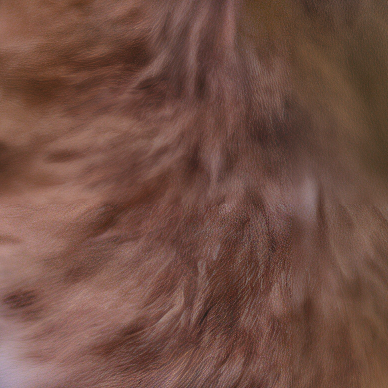} &
        \includegraphics[width=0.12\textwidth]{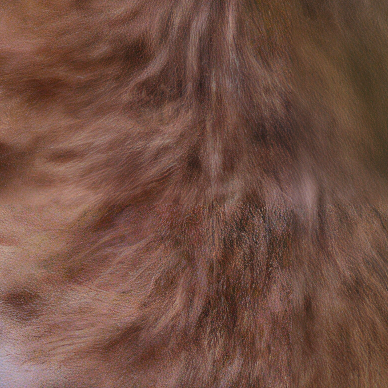} &
        \includegraphics[width=0.12\textwidth]{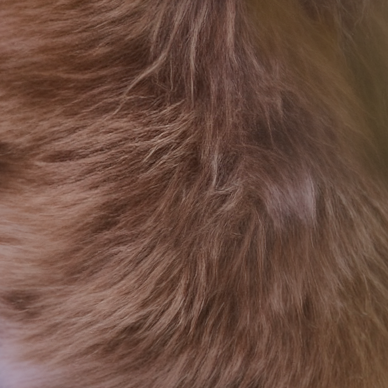} \\[-1mm]
        & \raisebox{-0.75\height}[0pt][0pt]{ (g) LDM-30} & \raisebox{-0.75\height}[0pt][0pt]{ (h) LDM-100} & \raisebox{-0.75\height}[0pt][0pt]{ (i) ResShift-15} & \raisebox{-0.75\height}[0pt][0pt]{ (j) SinSR} & \raisebox{-0.75\height}[0pt][0pt]{ (k) SAMSR} \\
    \end{tabular}
    \caption{Qualitative comparisons on real-world examples. Please zoom in for a better view.}
    \label{fig:qualitative_comparison}
    \vspace{-0.2cm}
\end{figure*}

\begin{table*}[h!] \small
    \centering
    \setlength{\tabcolsep}{20pt}  
    \renewcommand{\arraystretch}{1.1}  
    \small  

    \begin{tabular}{lcccc}
        \toprule
        \multirow{2}{*}{\textbf{Methods}} 
        & \multicolumn{2}{c}{\textbf{RealSR}} 
        & \multicolumn{2}{c}{\textbf{RealSet65}} \\
        \cmidrule(lr){2-3} \cmidrule(lr){4-5}
        & \textbf{CLIPIQA$\uparrow$} & \textbf{MUSIQ$\uparrow$} 
        & \textbf{CLIPIQA$\uparrow$} & \textbf{MUSIQ$\uparrow$} \\
        \midrule
        ESRGAN~\cite{wang2018esrgan}             & 0.2362         & 29.048         & 0.3739         & 42.369 \\
        RealSR-JPEG~\cite{ji2020real}        & 0.3615         & 36.076         & 0.5282         & 50.539 \\
        BSRGAN~\cite{zhang2021designing}             & 0.5439         &\underline{63.586}& 0.6163         & \textbf{65.582} \\
        SwinIR~\cite{liang2021swinir}             & 0.4654         & 59.636         & 0.5782         & 63.822 \\
        RealESRGAN~\cite{wang2021real}         & 0.4898         & 59.678         & 0.5995         & 63.220 \\
        DASR~\cite{liang2022efficient}               & 0.3629         & 45.825         & 0.4965         & 55.708 \\
        LDM-15~\cite{rombach2022high}             & 0.3836         & 49.317         & 0.4274         & 47.488 \\
        \midrule
        ResShift-15~\cite{yue2024resshift}        & 0.5958         & 59.873         & 0.6537         & 61.330 \\
        SinSR-1~\cite{wang2024sinsr}               & \underline{0.6887}& 61.582 & \underline{0.7150} & 62.169 \\
        SAMSR (Ours)                 & \textbf{0.7179} &\textbf{63.696} & \textbf{0.7324} & \underline{65.089} \\
        \bottomrule  
    \end{tabular}

    
    \caption{Quantitative results on two real-world datasets. 
    The best and second best results are highlighted in \textbf{bold} and \underline{underline}, respectively.}
    \vspace{-0.4cm}
    \label{tab:real_world_results}
\end{table*}

\begin{table*}[h] \small
    \centering
    \setlength{\tabcolsep}{16pt}  
    \renewcommand{\arraystretch}{1.1}  
    \small  
    \begin{tabular}{lccccc}
        \toprule
        \textbf{Method} & \textbf{PSNR$\uparrow$} & \textbf{SSIM$\uparrow$} & \textbf{LPIPS$\downarrow$} & \textbf{CLIPIQA$\uparrow$} & \textbf{MUSIQ$\uparrow$} \\
        \midrule
        ESRGAN~\cite{wang2018esrgan}           & 20.67 & 0.448 & 0.485 & 0.451 & 43.615 \\
        RealSR-JPEG~\cite{ji2020real}      & 23.11 & 0.591 & 0.326 & 0.537 & 46.981 \\
        BSRGAN~\cite{zhang2021designing}           & 24.42 & 0.659 & 0.259 & 0.581 & \textbf{54.697} \\
        SwinIR~\cite{liang2021swinir}           & 23.99 & 0.667 & 0.238 & 0.564 & 53.790 \\
        RealESRGAN~\cite{wang2021real}       & 24.04 & 0.665 & 0.254 & 0.523 & 52.538 \\
        DASR~\cite{liang2022efficient}             & 24.75 & \textbf{0.675} & 0.250 & 0.536 & 48.337 \\
        LDM-30~\cite{rombach2022high}           & 24.49 & 0.651 & 0.248 & 0.572 & 50.895 \\
        LDM-15~\cite{rombach2022high}           & \underline{24.89} & 0.670 & 0.269 & 0.512 & 46.419 \\
        \midrule
        ResShift-15~\cite{yue2024resshift}      & \textbf{24.90} & \underline{0.673} & 0.228 & 0.603 & 53.897 \\
        SinSR-1~\cite{wang2024sinsr}             & 24.56 & 0.657 & \underline{0.221} & \underline{0.611} & 53.357 \\
        SAMSR (Ours)               & 24.74 & 0.666 & \textbf{0.217} & \textbf{0.619} & \underline{54.146} \\
        \bottomrule
    \end{tabular}
    \caption{Quantitative results on \textit{ImageNet-Test}. The best and second best results are highlighted in \textbf{bold} and \underline{underline}, respectively.}
    \label{tab:imagenet_results}
    \vspace{-0.4cm}
\end{table*}

\noindent \textbf{Compared Methods.} We compare our method with several representative SR models, including RealSR-JPEG \cite{ji2020real}, ESRGAN \cite{wang2018esrgan}, BSRGAN \cite{zhang2021designing}, SwinIR \cite{liang2021swinir}, RealESRGAN \cite{wang2021real}, DASR \cite{liang2022efficient}, LDM \cite{rombach2022high}, ResShift \cite{yue2024resshift} and SinSR \cite{wang2024sinsr}. 
 
\noindent \textbf{Metrics.} To evaluate the fidelity of our method on synthetic datasets with reference images, we used PSNR, SSIM, and LPIPS \cite{zhang2018unreasonable}. Additionally, two recent non-reference metrics were employed to assess the realism of the generated images: CLIPIQA \cite{wang2023exploring}, which leverages a pretrained CLIP \cite{radford2021learning} model on a large-scale dataset, and MUSIQ \cite{ke2021musiq}.

\noindent \textbf{Training Details.} To ensure a fair comparison, we adopted the same experimental configuration and backbone architecture as described in prior work. However, our approach introduces modifications to the forward diffusion process and corresponding loss function. These adjustments enable a significant reduction in the number of training iterations compared to existing models. Specifically, our model achieves convergence in only 10,000-15,000 iterations. We attribute this improvement to the integration of a semantic consistency loss, which accelerates the convergence of the student model and further optimizes the training efficiency. 

\subsection{Experimental Results}

\noindent \textbf{Evaluation on Real-world Datasets.} We comprehensively evaluate SAMSR on both real-world and synthetic datasets to demonstrate its robustness and effectiveness across diverse scenarios. For real-world evaluation, we utilize RealSR \cite{cai2019toward} and RealSet65 \cite{yue2024resshift}. Both datasets exhibit diverse degradation patterns and lack ground truth. SAMSR is compared against SOTA SR methods, using non-reference metrics CLIPIQA and MUSIQ \cite{wang2023exploring,ke2021musiq}. As shown in Table \ref{tab:real_world_results}, SAMSR achieves superior performance in both metrics, benefiting from its semantic-guided noise sampling and region-aware diffusion dynamics.

\begin{table*}[t!] \small
    \centering
    \setlength{\tabcolsep}{8pt} 
    \renewcommand{\arraystretch}{1.1} 

    \begin{tabular} {p{0.05\textwidth}p{0.05\textwidth}p{0.05\textwidth}p{0.1\textwidth}p{0.1\textwidth}p{0.1\textwidth}p{0.1\textwidth}}
        \toprule
        \multicolumn{3}{c}{\textbf{Hyper-parameters}} & \multicolumn{2}{c}{\textbf{RealSR}} & \multicolumn{2}{c}{\textbf{RealSet65}} \\
        \cmidrule(lr){1-3} \cmidrule(lr){4-5} \cmidrule(lr){6-7}
        \textbf{$m$} & \textbf{$p$} & \textbf{$\kappa$} 
        & \textbf{CLIPIQA$\uparrow$} & \textbf{MUSIQ$\uparrow$} 
        & \textbf{CLIPIQA$\uparrow$} & \textbf{MUSIQ$\uparrow$} \\
        \midrule
        1/2 & 0.3 & 2.0 
        & 0.6992 & 60.483 & 0.7193 & 62.451 \\
        1/4 & 0.3 & 2.0 
        & 0.7019 & 61.642 & 0.7221 & 62.633 \\
        \textbf{1/5} & \textbf{0.3} & \textbf{2.0} 
        & \textbf{0.7179} & \textbf{63.696} & \textbf{0.7324} & \textbf{65.089} \\
        1/6 & 0.3 & 2.0 
        & 0.7092 & 62.734 & 0.7251 & 62.914 \\
        1/8 & 0.3 & 2.0 
        & 0.7119 & 62.385 & 0.7291 & 64.218 \\
        1/10 & 0.3 & 2.0 
        & 0.7069 & 61.982 & 0.7216 & 63.814 \\
        1/20 & 0.3 & 2.0 
        & 0.6953 & 61.492 & 0.7194 & 62.843 \\
    \bottomrule
    \end{tabular}
    \caption{Quantitative results of models under different Hyper-parameters ($m$, $p$, $\kappa$).}
\vspace{-0.4cm}
    \label{tab:Hyper-parameters}
\end{table*}

\noindent \textbf{Evaluation on Synthetic Datasets.} For synthetic datasets, we follow the standard pipeline to create LR inputs from 3000 HR images randomly selected from ImageNet \cite{wang2024sinsr}. Evaluation metrics include fidelity measures (PSNR, SSIM, LPIPS) \cite{zhang2018unreasonable} and perceptual quality metrics (CLIPIQA, MUSIQ) \cite{wang2023exploring,ke2021musiq}. Results in Table \ref{tab:imagenet_results} indicate that SAMSR attains comparable fidelity metrics to existing diffusion-based models while significantly improving perceptual quality. The introduction of semantic masks enables pixel-wise adjustment of residual transfer rates and noise strengths, enhancing detail preservation in semantically rich regions and allowing SAMSR to outperform SinSR in balancing detail recovery and perceptual realism. These comprehensive evaluations demonstrate SAMSR’s effectiveness and versatility in both real-world and controlled synthetic environments.

\subsection{Model Analysis}

\textbf{Hyper-parameter $m$.} The hyper-parameter \( m \) controls the noise addition speed and intensity for pixels with different levels of semantic richness during the forward diffusion process. Table~\ref{tab:Hyper-parameters} summarizes the performance of SAMSR on the Realset65 and RealSR dataset under different values of \( m \). We observe that both excessively large and small values of \( m \) degrade the model's authenticity. Experiments show that when \( m \) is within the range of [1/5, 1/8], our method achieves outstanding results on both the Realset65 and RealSR datasets. Therefore, in this paper, we set \( m \) to 1/5.

\begin{table}[h] \small
\centering
\begin{tabular}{lcc}
\toprule
\textbf{Method} & \textbf{CLIPIQA$\uparrow$} & \textbf{MUSIQ$\uparrow$} \\
\midrule
\textbf{SAMSR($\eta_t^{\text{new}}, \kappa$)} & 0.7208 & 63.613 \\
\textbf{SAMSR($\eta_t, \kappa^{\text{new}}$)} & 0.7194 & 63.241 \\
\textbf{SAMSR($\eta_t^{\text{new}}, \kappa^{\text{new}}$)} & \textbf{0.7324} & \textbf{65.089} \\
\bottomrule
\end{tabular}
\caption{A comparison of the SAMSR method with different hyper-parameters (evaluation on RealSet65 datasets).}
\label{tab:eta_kappa}
\vspace{-0.4cm}
\end{table}

\noindent \textbf{Hyper-parameters \( k \) and \( \eta_t \).} \( \eta_t \) is the residual transition ratio defined during the diffusion process, which controls the gradual transition speed from the HR image to the LR image in the Markov chain. \( \kappa \) is the control parameter for noise intensity during the diffusion process, affecting the strength of noise introduced at each diffusion step. In this paper, we modify \( \eta_t \) and \( \kappa \) using the hyper-parameter \( m \), incorporating dense semantic guidance. This ensures that semantically significant regions are prioritized for finer recovery during a single step of reverse diffusion, while background regions maintain higher global consistency. Table~\ref{tab:eta_kappa} further explores the effects of individually modifying \( \eta_t \) and \( \kappa \).

\noindent \textbf{Effective of Pixel-wise Sampling.} Previous research has shown that learning the deterministic mapping between \( x_t \) and \( x_0 \) is hindered by the non-causal nature of the generative process. However, our experiments demonstrate that modifying the local noise intensity across different masked regions of an image effectively mitigates this issue. This allows the student network in the knowledge distillation process to better solve the ODE process in a single step, while maintaining the same model size. 

\begin{table}[t] \small
\centering
			\resizebox{1\linewidth}{!}{%
\begin{tabular}{lcccc}
\toprule
\textbf{} & \textbf{Num. of Iters} & \textbf{Train. Time} & \textbf{CLIPIQA$\uparrow$} & \textbf{MUSIQ$\uparrow$} \\
\midrule
ResShift & 500k &~7.64days & 0.6537 &61.330 \\
SinSR & 30k &~2.57days & 0.7150  & 62.169 \\
\textbf{SAMSR} & \textbf{10-15k} &\textbf{~1.89days} & \textbf{0.7324} & \textbf{65.089} \\
\bottomrule
\end{tabular}}
\caption{A comparison of the training time cost and results on NVIDIA RTX4090.}
\vspace{-0.4cm}
\label{train}
\end{table}

\noindent \textbf{Effective of Consistency Semantic Loss.} Our consistency semantic loss aims to improve the model's ability to understand and apply semantic information. Specifically, we establish a loss function based on the semantic weight differences between the ground-truth image and the predicted image \( x_0 \), enhancing the model’s ability to recover semantic details. From Table~\ref{train}, we observe that the consistency semantic loss not only significantly accelerates the model’s convergence speed but also effectively improves its performance, demonstrating its effectiveness during the training process.

\section{Conclusion}
\label{sec:conclusion}

In this paper, we propose a semantic segmentation-guided diffusion model named SAMSR. Specifically, we introduce a pixel-wise sampling framework based on semantic segmentation, where noise is sampled within masked regions to retain the spatial and semantic characteristics of the original image. Additionally, we leverage segmentation masks to derive pixel-level sampling hyperparameters, enabling differentiated noise schedules for pixels with varying semantic richness. This ensures that semantically rich regions achieve significant recovery within a single sampling step. Furthermore, we propose a semantic consistency loss to accelerate the convergence of the model. Experimental results show that our approach achieves significant performance improvements, particularly for SR tasks on semantically complex images.

\bibliographystyle{named}
\bibliography{ijcai25}

\end{document}